\documentclass[letterpaper, 10 pt, conference]{ieeeconf}  % Comment this line out if you need a4paper

\usepackage{booktabs}
\usepackage{amsmath,amsfonts}
\usepackage{algorithmic}
\usepackage{multirow}
\usepackage{makecell}
\usepackage{amsmath}
\usepackage{algorithm}
\usepackage{array}
\usepackage[caption=false,font=normalsize,labelfont=sf,textfont=sf]{subfig}
\usepackage[capitalise]{cleveref}
\usepackage{textcomp}
\usepackage{stfloats}
\usepackage{url}
\usepackage{verbatim}
\usepackage{graphicx}
\usepackage{cite}
\usepackage{xcolor}
\IEEEoverridecommandlockouts                              % This command is only needed if 
\title{\LARGE \bf
Rotate Disks to Reach Farther: Design and Modeling of a Novel Reconfigurable Tendon Driven Manipulator 
}

\author{
Sabyasachi Dash$^{1}$, Yangkun Liu$^{2}$, Will Hunter$^{1}$, John Golden$^{2}$, Girish Krishnan$^{1}$%
\thanks{$^{1}$Department of Industrial \& Enterprise Systems Engineering, University of Illinois Urbana-Champaign}%
\thanks{$^{2}$Department of Mechanical Science \& Engineering, University of Illinois Urbana-Champaign}%
}
\begin{document}

\maketitle
\thispagestyle{empty}
\pagestyle{empty}

%%%%%%%%%%%%%%%%%%%%%%%%%%%%%%%%%%%%%%%%%%%%%%%%%%%%%%%%%%%%%%%%%%%%%%%%%%%%%%%%

\begin{abstract}
Rerouting the tendon path in tendon driven continuum manipulators (TDCMs) enables a broad range of deformation modes. This work presents a Reconfigurable TDCM design which allows independent rotation of intermediate spacer disks, thereby locally rerouting the tendon and achieving non-trivial backbone spatial deformations. Two such designs, (a) Manual Disk Locked (MDL) and (b) Continuous Disk Rotor (CDR) manipulators are presented to achieve disk rotations before and during operation, respectively. A predictive static model based on the piecewise constant strain (PCS) assumption is developed within a potential energy minimization framework,  incorporating (a) disk rotations, (b) discrete tendon paths between disk segments, (c) rigid thickness of spacer disks, and (d) elasticity of the tendons. The model is validated against experimental results, demonstrating an average tip error of $1.2\%$ of the manipulator's total length for parallel tendon routing and around $3\%$ for the case when multiple disks are rotated. The computation time is an order of magnitude lower than the state of the art Cosserat rod solver.  
% In tendon driven continuum manipulators (TDCMs), reconfiguring the tendon routing enables tailored spatial deformation of the backbone. This work presents a design in which tendons can be rerouted either prior to or after actuation by actively rotating the individual spacer disks. Each disk rotation thus adds a degree of freedom to the actuation space, complicating the mapping from a desired backbone curve to the corresponding actuator inputs. However, when the backbone shape is projected into an intermediate space defined by curvature and torsion (C-T), patterns emerge that highlight which disks are most influential in achieving a global shape. This insight enables a simplified, sequential shape-matching strategy: first, the proximal and intermediate disks are rotated to approximate the global shape; then, the distal disks are adjusted to fine-tune the end-effector position with minimal impact on the overall shape. The proposed actuation framework offers a model-free alternative to conventional control approaches, bypassing the complexities of modeling reconfigurable TDCMs.
\end{abstract}
%
% \begin{IEEEkeywords}
% Continuum manipulators, Reconfigurable Tendon Routing, Shape Matching 
% % Continuum robots, End Effectors, Shape Control, Soft robotics, Servomotors, Root mean square, Tendons, Shape, Kinematics, Bending
% \end{IEEEkeywords}
\section{Introduction}
Tendon driven continuum manipulators (TDCMs) \cite{Walker2013} represent a prominent and widely explored category of soft continuum robots, which feature a slender, flexible backbone actuated using cables/tendons. Their inherent compliance and high dexterity enables them to maneuver through highly constrained environments, making them suitable for minimally invasive surgeries \cite{burgner-kahrs2015}, industrial applications (e.g. in-situ defect inspection \cite{10802783}), and grippers in space \cite{10801537}. In TDCMs, it is well known that the tendon routing along the backbone governs the deformation modes, and in turn the shape of the manipulator. In conventional TDCMs, the tendons are routed parallel to the backbone, constraining the feasible deformation modes/shape profiles. Moreover, with parallel tendons, the reachable workspace is constrained to a semi-ellipsoid \cite{jessicahelical}. Generalized tendon routings with  helical or polynomial paths have shown advantages, but are still limited by the range spanned by the deformation modes of individual tendons. This makes it more difficult to emulate examples in nature such as octopus tentacles and elephant trunks that can demonstrate multi-axial bending, spiraling, or even multiple curvatures.  
\begin{figure}[h]
    \centering        \includegraphics[width=0.43\textwidth]{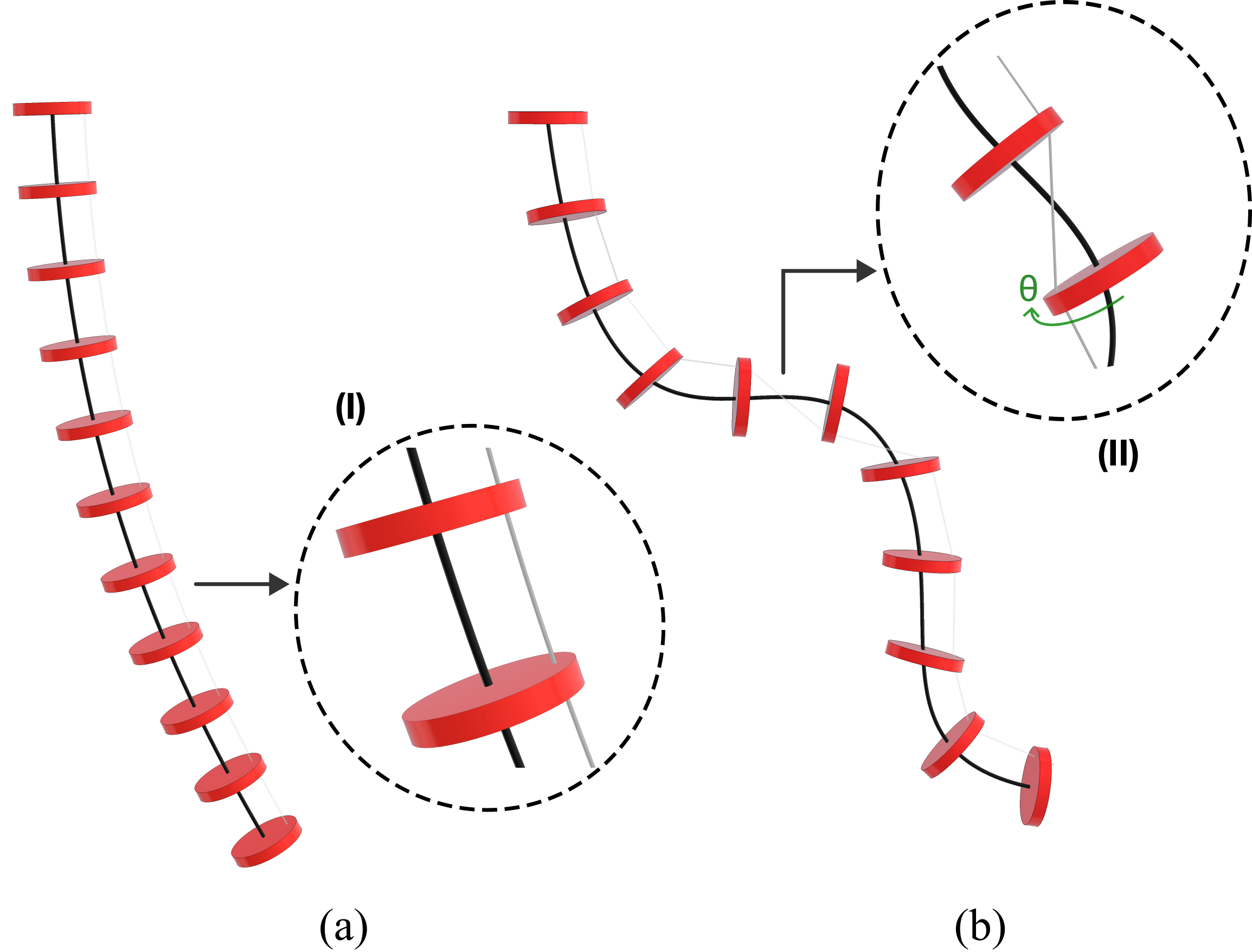} 
    \caption{Conventional TDCMs (a) rely on fixed predefined parallel tendon routing which constrains their attainable deformation modes. The proposed RTDCM design (b) leverages intermediate spacer disk rotation to locally reroute the tendon allowing more versatile backbone spatial configurations. Though the backbone deforms as a smooth continuous curve, the tendon follows a discrete linear path between the rotated disk segments (II). }
     \label{abs}
\end{figure}
In this paper, we present the design of a novel reconfigurable TDCM (RTDCM) whose tendon routing can be reconfigured based on the deformation requirements of a given task. The tendon routings are changed by relative rotation of spacer disks about the backbone's axis. 
% Spacer disks are used to route the tendons along the backbone length and maintain a specific geometric relationship between the tendon and the backbone. 
To realize this concept physically, we demonstrate \textit{two} RTDCM prototypes: (a) the MDL (Manual Disk Locking) manipulator, where the disks are rotated and locked before tendon actuation, and (b) the CDR (Continuous Disk Rotor) manipulator, capable of rotating the spacer disks during operation. \cref{abs} demonstrates the design concept, where each intermediate spacer disk used to guide the tendon can be rotated and locked in a desired configuration, enabling on-demand tendon path reconfiguration.

The optimal design and eventual control of the manipulators require a predictive model that is accurate, yet computationally efficient. Though TDCM modeling, including generalized tendon routings, is widely studied, a feasible model requires incorporating additional features unique to RTDCM designs such as disk rotation and its relation to the tendon path, discrete tendon routing, tendon elasticity, and the effect of spacer disk thickness. In this work, we propose an energy minimization framework using piecewise constant strain (PCS) kinematics that incorporates these features. 
The contributions of this paper are three-fold: \\
(1) First, we demonstrate the designs of two novel manipulators, where intermediate spacer disks can be rotated to create a local rerouting of the tendon. The disks can be rotated manually and locked at a desired angle in the MDL manipulator, or can be actively rotated using servo motors in the CDR design.
(2) Second, we develop a potential energy minimization framework with a Piecewise Constant Strain (PCS) assumption to solve for static equilibrium of the manipulator, taking into consideration: (a) discrete straight tendon segments between locally rotated disks, (b) rigid thickness induced by the spacer disks, and (c) elastic properties of the tendon(s). 
Although most modeling techniques for TDCMs involve solving the equilibrium equations in the strong form, variational formulations based on potential energy minimization for statics and Lagrangian energy formulations for dynamics have also been explored [6]. We envision that an energy-based formulation will aid future work involving design optimization and control. This model is further validated against a higher-fidelity Cosserat Rod mechanics-based solver to benchmark its performance.
(3) Lastly, we design physical prototypes for the two proposed RTDCMs and validate the MDL manipulator with the developed model. The errors observed comply with the standard error tolerances reported in TDCM literature.

\section{Related Work}
\subsection{Non-parallel Tendon Routings}
% In conventional TDCMs, the tendons are routed parallel to the backbone, constraining the feasible deformation modes/ shape profiles. Moreover, with parallel tendons the reachable workspace is constrained to a semi-ellipsoid \cite{jessicahelical}. To mitigate this, non-straight tendon routings have shown to be beneficial in expanding the robot's workspace. 
General tendon routings were explored in \cite{rucker2011}, and a prototype was demonstrated with helical and polynomial routings. \cite{jessicahelical} reported the increase in reachable workspace by $\sim$400\% upon addition of a helically routed tendon to a TDCM with three parallel tendons. In HelixFlex \cite{gerboni2015}, 12 of the 18 tendons followed non-straight routes, achieving a variety of 3D shapes. A cross-helical tendon was proposed in \cite{7353643} which was capable of generating S-shapes within a plane. Though most of these manipulators showcased designs with non-straight tendons, the routing was predefined, without having the capabililty to change it based on task-specific requirements. A more modular design was proposed in \cite{grassmann2022fas} where the extrinsic rotation of the backbone enabled twist in the entire manipulator. However, the intermediate spacer disks were free-floating and not capable of being controlled independently.
Our proposed RTDCMs address these issues by allowing independent rotation of each disk, making more non-trivial tendon pathways feasible.
\subsection{Modeling methods}
Kineto-static models for TDCMs have been thoroughly explored, most of which are documented in \cite{rao2021model}. The models range from simplified constant curvature to higher fidelity models predominantly implementing Cosserat Rod Theory \cite{antman2005nonlinear}. Dynamics modeling methods have also been explored, with general soft robotic modeling platforms including \cite{9895355} showing a certain degree of success in modeling conventional TDCMs. 
Though the majority of modeling frameworks are implemented to parallel/straight routed tendons, the model presented in \cite{rucker2011} proposed a Cosserat Rod theory (CRT) based framework for general tendon routings. Though the tendon paths demonstrated in \cite{rucker2011} were defined as functions of their radial distance from the backbone centerline, provisions for discrete jumps in the tendon path were discussed through disk-level boundary conditions. \cite{rao2021model} demonstrated the effect of discrete tendon routings between segments, analyzing it as a partially constrained tendon path. Although CRT based solvers show increased accuracy, they usually rely on computationally expensive solutions to a set of ODEs or PDEs (for dynamics), which limit their real-time applications particularly for simulation and control \cite{muhmann2025toward}. This makes other simplified models predominant in TDCM modeling. Piecewise Constant curvature (PCC) models, which assume the backbone curvature to be constant between disk segments, have been used extensively (\cite{1588999}, \cite{webster2010design}) and have demonstrated sufficient success in controls \cite{10847881} due to their computational advantage. Recently, \cite{10606060} implemented PCC in a helical routed model showcasing effects of specific helical paths in reducing computational loads. To take into account the out-of-plane bending (torsion), Piecewise Constant Strain (PCS) models have been developed \cite{8500341}, which assume constant magnitude of material strain vectors in discretized rod segments. The PCS model has demonstrated better accuracy when compared with PCC models, and computationally less expensive than Cosserat models (\cite{8500341}, \cite{shamilyan2023intelligence}).
Most of the non-straight routing approaches either implement a predefined path, consider only helical routing, or do not consider individual tendon segments to be linear. 

In addition to selecting a particular modeling approach for the flexible backbone, studies have demonstrated that the tendon's material stiffness affects the actuation \cite{tendon-elasticity2}. In addition, we note that the rigid spacer disks attached to the backbone affect its overall bending length, which needs to be incorporated in the modeling framework. Although these effects are modeled independently in different manipulator designs using different mechanical models, to our knowledge, no modeling framework exists that takes all of these effects into account while considering reconfigurable tendon routing. Towards this, we first present the design methodology for our novel RTDCM, followed by the detailed framework of the proposed PCS-based energy formulation.

\section{Design}
\label{design section}
\label{design}
\begin{figure}[h]
    \centering    \includegraphics[width=0.4\textwidth]{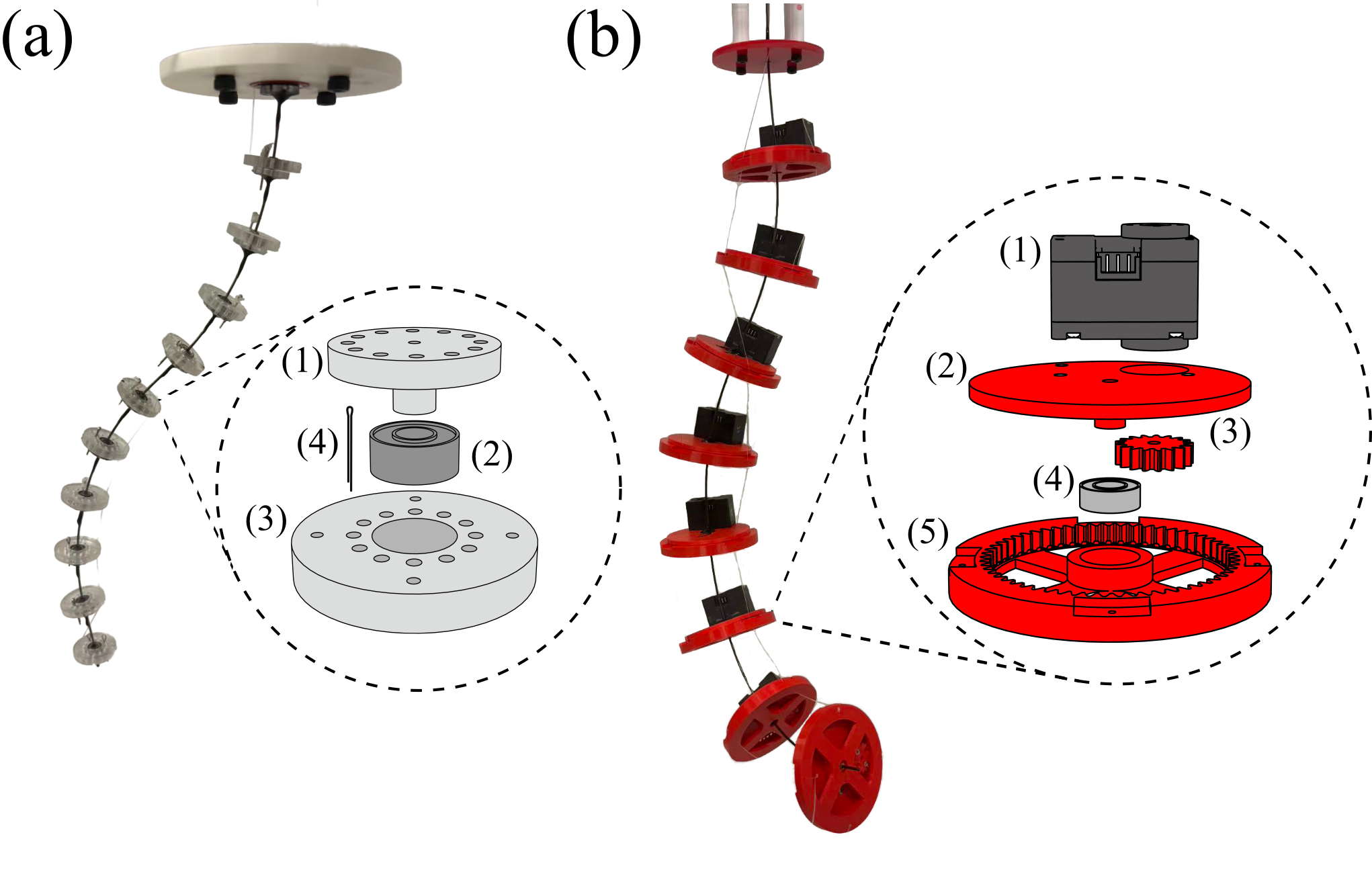} 
    \caption{
    The spacer disk for the MDL manipulator (a) is split into two main components, with the top bushing (1) attached to the backbone using epoxy and the bottom disk (3) freely rotating on the bearing (2) to locally reroute the tendon. A desired tendon configuration is locked by inserting the cotter pin (4). The CDR manipulator (b) enables active tendon rerouting using the servo motor (1), which sits on the larger top bushing (2) attached to the backbone and is meshed with the rotating disk (5) via the spur gear (3).}
    % (a) Small manipulator design with each disk assembly capable of rotating the tendon locally. Upper disc (1) bonded to the spine via epoxy resin, is connected to the rotating disk (3) through a bearing (2). The lower disc (3) is manually rotated to the target orientation and secured using the cotter pin (4).(b) Big manipulator design with each disk assembly capable of rotating the tendon locally. The servo motor (1) sits on the fixed mount (2) which is attached to the backbone using epoxy. The fixed mount is connected to the rotating disk (5) through a bearing (4), and a gear (3) attached to the servo motor-head rotates (5) which holds the tendon. (c) Experimental setup with stepper motor (1) to actuate the tendon, and camera1 (2) camera2 (3) to retrieve track coordinates of the backbone. (4) is the manipultor under validation tests.}
    \label{fig:design_modular}
\end{figure}
This section describes the design and fabrication procedures for both the MDL and the CDR manipulators. The flexible backbone of each is made out of Nitinol (NiTi), a nickel-titanium alloy, with multiple spacer disks evenly distributed along its length. Compared to a conventional TDCM, the spacer disks for both manipulator prototypes are modified to enable tendon rerouting, as demonstrated in \cref{fig:design_modular}. The design parameters for both models are tabulated in \cref{tab:robot_params}. For the MDL manipulator \cref{fig:design_modular}(a), each spacer disk is made of two split-parts, where the top bushing (1) is fixed to the 0.5 mm radius NiTi backbone using an epoxy-based adhesive and is connected to the bottom rotating disk (3) through a 7 mm OD bearing (2). Up to four tendons can be routed through the outer holes on the bottom rotating disk, although only one is used for the experiments carried out in this paper. An inner set of holes on both the top (1) and bottom (3) part, spaced $30^\circ$ apart from each other, allows for manual disk rotation, which can be locked into a rerouted tendon configuration using cotter pins (4). To ensure precise hole diameters, the disk components were printed using clear photopolymer resin in a Formlabs Form-4 3D printer. The spacer disks, along with the adhesive, can alter the overall stiffness and flexible bending length of the backbone, which necessitate parameter calibration as will be discussed in \cref{calib}. Nylon monofilament fishing lines are used as tendons, which are susceptible to axial elongation, and thus need to be considered in the statics model. This simple MDL manipulator is lightweight and is therefore considered for experimental testing and model validation in this work. 
\begin{table}[htbp]
\caption{Manipulator Design Parameters}
\label{tab:robot_params}
\centering
\setlength{\tabcolsep}{8pt} 
\begin{tabular}{lcccc}
\toprule
\textbf{Parameter} & \textbf{Symbol} & \textbf{MDL} & \textbf{CDR} & \textbf{Unit} \\
\midrule
% Total Length & $L_0$ & 552 & 270 & mm \\
Number of segments & $n$ & 10 & 8 & -- \\
Segment length & $L_{\mathrm{seg}}$ & 27 & 69 & mm \\
Backbone diameter & $d_b$ & 1.0 & 1.5 & mm \\
Backbone density & $\rho$ & 6450 & 6450 & kg/m$^3$ \\
Tendon Young's modulus & $E_t$ & 4.5 & 4.5 & GPa \\
Tendon radial offset & $R_{\mathrm{rad}}$ & 8 & 34 & mm \\
Disk radius & $R_{\mathrm{disk}}$ & 10 & 40 & mm \\
Single disk mass & $m$ & 2 & 44 & g \\
\bottomrule
\end{tabular}
\end{table}
In order to actively rotate the disks during operation, we present the CDR design, where the rotation of each disk can be controlled extrinsically using lightweight servo motors as demonstrated in \cref{fig:design_modular}(b). Each
spacer disk assembly is now an independently actuated module
and consists of five parts: a top bushing mount, continuous
servo motor, internal ring gear, spur gear, and central bearing
\cref{fig:design_modular}(b). The top bushing (2) houses the continuous servo motor (DYNAMIXEL XL330-M288-T, ROBOTIS Co. LTD, Gangseo-gu, Seoul, Korea) and is rigidly fixed to the NiTi backbone. The servo motor features an absolute multi-turn encoder which tracks the angular position of the tendon. The servo motors are daisy chained together in two groups to prevent current overload and powered via a U2D2 powerhub. Communication is done with a single shared bus operated by an OpenRB-150 embedded controller. A seventeen-tooth spur gear (3) is rigidly fixed to the head of the servo horn and meshed with a fifty-eight-tooth internal ring gear (5), yielding a gear ratio of 3.41. The fixed mount (2) is connected to the rotating disk (5) concentrically via a bearing (4) and a spur gear (3) meshed between them. Upon actuation of the servo, the
rotating disk (5) alters the tendon’s angle, thereby locally rerouting the tendon within the segment between the disks above and below it. This automatic disk rotation enables tendon rerouting both prior to and during operation.

\section{Modeling}
\label{modelsection}
% \note{\textbf{GK: suggested narrative. Feel free to include this in the introduction or related work where it may make more sense}
% A predictive model for the deformed backbone shape of the proposed TDCM, expressed as a function of tendon displacement or tension, must incorporate several key features:(a) Disk rotation: the tendon path must be modeled as a function of one or more disk rotations; (b) Discrete tendon path: the sizable spacing between disks may cause tendon segments to remain straight between them rather than follow a continuous curve; (c) Effective bending length: the finite thickness of the spacer disks and the epoxy used to rigidly bond them to the backbone reduces the portion of the backbone that participates in bending; (d) Tendon elasticity: elastic energy stored in the tendon violates the inextensibility assumption and results in reduced overall backbone\textbf{ \cite{9789127}}. Although these effects have been considered individually in previous models\textbf{\cite{tenden-elasticity1}\cite{tendon-elasticity2}}, no existing framework incorporates all of them in a way suitable for reconfigurable tendon routing. }
This section details the modeling framework based on solving the equilibrium from a minimum potential energy formulation, considering Piecewise Constant Strain (PCS) for each segment. First, a single backbone segment considering local disk rotations and rigid spacer disk thickness is solved analytically, and the desired Jacobian sensitivities are propagated for the entire $n$ - segment manipulator. Then the static equilibrium is obtained by solving the optimality conditions of the total potential energy. Finally, we validate the PCS model with a higher fidelity Cosserat ODE framework as discussed in \cite{rucker2011}, \cite{rao2021model}.

\subsection{Piecewise Constant Strain: Single segment solution}
\label{singlesegment}
\begin{figure}[h]
    \centering    
    \includegraphics[width=0.37\textwidth]{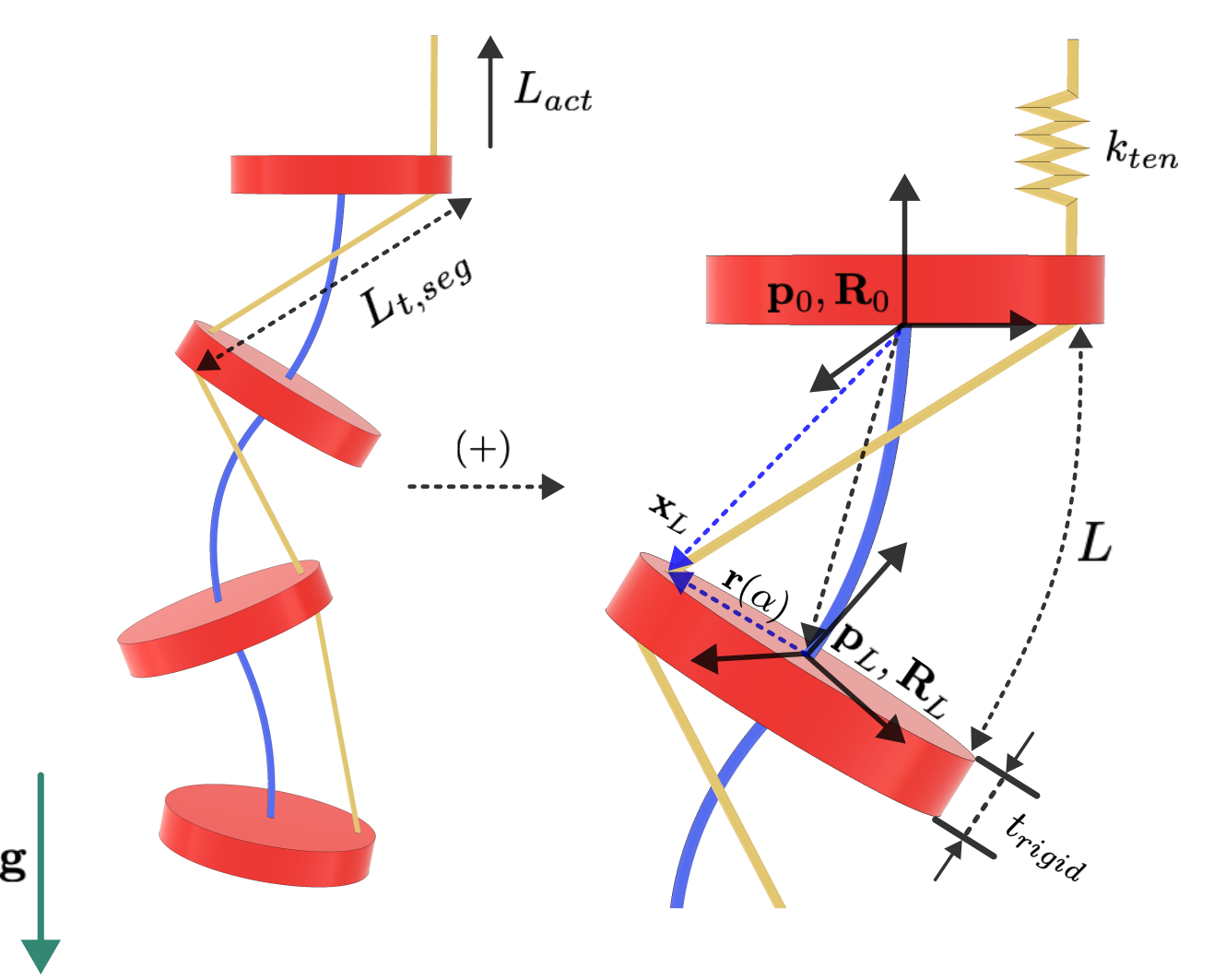} 
    \caption{Representative manipulator description (left) with piecewise constant strain segment (right) illustrating the various parameters and notations used in the model. $\mathbf{p_0}$ and $\mathbf{R_0}$ correspond to the base frame pose.}
    \label{model}
\end{figure}
The backbone of the manipulator is modeled as a Kirchhoff rod, neglecting stretch and shear effects. The kinematics of the centerline in space can be described by \cite{antman2005nonlinear}:
\begin{equation}
\label{eq:cosserat_kinematics}
\mathbf{p}'(s)=\mathbf{R}(s)\mathbf{v}(s),
\qquad
\mathbf{R}'(s)=\mathbf{R}(s)\,\mathbf{u}(s)^\wedge
\end{equation}
where $(\cdot)'=\partial(\cdot)/\partial s$, $\mathbf{v}(s)$ is the local translational strain and $\mathbf{u}(s)$ is the local rotational strain. The hat operator maps $\mathbf{u}=[u_1\;u_2\;u_3]^{\top}$ to the skew-symmetric matrix $\mathbf{u}^\wedge$ such that $\mathbf{u}^\wedge\mathbf{x}=\mathbf{u}\times\mathbf{x}$.

Consider a single manipulator segment as illustrated in \cref{model}. The effective bending length is defined as $L = L_{\text{seg}} - t_{\text{rigid}}$  where $L_{\text{seg}}$ is the total segment length and $t_{\text{rigid}}$ accounts for the combined rigid thickness of the spacer disk and epoxy layer. Within each flexible section, the strain variables are assumed to be constant, $\mathbf{u}(s)\equiv\mathbf{u}=[k_1\;k_2\;k_3]^{\top}$, where $k_1,k2$ denote the planar curvatures and $k_3$ corresponds to the out of plane torsion, and $\mathbf{v}(s)\equiv\mathbf{v}=[0\;0\;1]^{\top}$, corresponding to an inextensible and un-shearable backbone. Let $\omega=\|\mathbf{u}\|$. Under the constant-strain assumption, the rotational equation in \eqref{eq:cosserat_kinematics} attains the closed-form solution $\mathbf{R}(L)=\mathbf{R}_0 \exp(\mathbf{u}^\wedge L)$. Using Rodrigues’ formula for the matrix exponential on $\mathrm{SO}(3)$, this becomes:
\[
\mathbf{R}(L)=\mathbf{R}_0
\left(
\mathbf{I}_3
+
\frac{\sin(\omega L)}{\omega}\mathbf{u}^\wedge
+
\frac{1-\cos(\omega L)}{\omega^2}(\mathbf{u}^\wedge)^2
\right).
\]
Substituting this exponential form into the translational equation in \eqref{eq:cosserat_kinematics} yields 
\[
\mathbf{p}(L)=\mathbf{p}_0+\mathbf{R}_0\,\mathbf{s}(L),
\]
\[
\mathbf{s}(L)
=
\left(
a_1\mathbf{I}_3
+
a_2\,\mathbf{u}^\wedge
+
a_3\,(\mathbf{u}^\wedge)^2
\right)\mathbf{v},
\]
with $a_1=L$, $a_2=(1-\cos(\omega L))/\omega^2$, and $a_3=L/\omega^2-\sin(\omega L)/\omega^3$. 
The rigid spacer, along with the epoxy adhesive, is modeled as a pure translation of length $t_{\text{rigid}}$ along the end tangent direction $\mathbf{R}(L)\mathbf{v}$ without additional rotation. The pose at the end of the complete segment can be formulated as:
\begin{equation}
\label{eq:segment_pose_with_thickness}
\mathbf{p}_{\text{out}}
=
\mathbf{p}(L)
+
t_{\text{rigid}}\,\mathbf{R}(L)\mathbf{v},
\qquad
\mathbf{R}_{\text{out}}
=
\mathbf{R}(L).
\end{equation}
% This provides the analytical solution to a single constant strain segment as a function of the rotational and linear strain vectors. 
The tendon between the two spacer disks attains a strictly linear path, which can be captured by the Euclidean distance between the tendon points. Following the $j^\text{th}$ segment, let the pose be given as $\mathbf{p}_j$ and $\mathbf{R}_j$. The disk $j$, when rotated by an angle $\alpha$ in the local frame, corresponds to a tendon attachment point given by $\mathbf{r}(\alpha_j) = [R_{\text{rad}}\cos\alpha_j\;\;R_{\text{rad}}\sin\alpha_j\;\;0]^\top$, where $R_{\text{rad}}$ corresponds to the tendon attachment point's distance from the centerline. Thus, the position vector of the tendon attachment point is: $\mathbf{x}_j = \mathbf{p}_j + \mathbf{R}_j \mathbf{r}(\alpha_j)$. 

\subsection{Multi–Segment Propagation}
\label{multiseg}
Consider the manipulator to consist of $n$ constant strain segments. Let the strain vector of segment $i$ be $\mathbf{u}_i \in \mathbb{R}^3$, and the global strain parameter vector $\mathbf{q} = [\,\mathbf{u}_1^\top \; \mathbf{u}_2^\top \; \cdots \; \mathbf{u}_n^\top\,]^\top \in \mathbb{R}^{3n}$. For each segment, the pose can be propagated to the next by: $\mathbf{p}_{i+1} = \mathbf{p}_i + \mathbf{R}_i \mathbf{d}_i$, 
$\mathbf{R}_{i+1} = \mathbf{R}_i \mathbf{R}_{\exp,i}$, 
with $\mathbf{R}_{\exp,i} = \exp(\mathbf{u}_i^\wedge L_i)$ and 
$\mathbf{d}_i = \mathbf{s}_i(\mathbf{u}_i) + t_{\text{rigid}} \mathbf{R}_{\exp,i} \mathbf{v}$. A potential energy based formulation will require gradient computations for obtaining static equilibrium solutions, which become numerically expensive when solving for multiple segments. Towards this, we propagate the Jacobians (sensitivities) with respect to the global strain vector $\mathbf{q}$. 
Let $\mathbf{P}_i = \partial \mathbf{p}_i / \partial \mathbf{q}$ and 
$\mathcal{R}_i = \partial \mathbf{R}_i / \partial \mathbf{q}$. Differentiating the segment recursion with respect to $\mathbf{q}$ yields the recursive relations 
\begin{equation}
\label{sensitivity}
\begin{aligned}
\mathcal{R}_{i+1} &= \mathcal{R}_i \mathbf{R}_{\exp,i}
+ \mathbf{R}_i \frac{\partial \mathbf{R}_{\exp,i}}{\partial \mathbf{q}}, \\
\mathbf{P}_{i+1} &= \mathbf{P}_i
+ \mathcal{R}_i \mathbf{d}_i
+ \mathbf{R}_i \frac{\partial \mathbf{d}_i}{\partial \mathbf{q}}.
\end{aligned}
\end{equation}
Since $\mathbf{R}_{\exp,i}$ and $\mathbf{d}_i$ depend only on the local strain $\mathbf{u}_i$, their derivatives with respect to the global parameter vector $\mathbf{q}$ vanish for all components not associated with segment $i$. Thus, the corresponding Jacobian contributions are nonzero only in the three columns of $\mathbf{q}$ corresponding to $\mathbf{u}_i$. 
This method of sensitivity propagation provides an $O(n)$ procedure for assembling the full manipulator Jacobian from single–segment analytic derivatives, without re-differentiating the entire kinematic chain.

\subsection{Total Potential Energy formulation}
The static equilibrium configuration is obtained by minimizing the total potential energy $\Pi(\mathbf{q})$, expressed as the sum of elastic potential energy of the backbone, as well as gravitational and tendon energy contributions. The manipulator is displacement-controlled, and the total tendon length is therefore enforced to be equal to the sum of the actuated length and tendon's extension. The gravitational energy terms are comprised of both the contributions from the spacer disks and the distributed weight of the backbone. 
\subsubsection{Elastic Potential Energy of the backbone}
With constant strain $\mathbf{u}_i$ over segment $i$ of length $L_i$, the
discretized bending energy is
\begin{equation}
\label{eq:bending_energy}
\Pi_{\text{EP}}(\mathbf{q})
=
\sum_{i=1}^n \frac{1}{2}\,L_i\,\mathbf{u}_i^\top \mathbf{K}\,\mathbf{u}_i,
\end{equation}
where $\mathbf{K}\in\mathbb{R}^{3\times 3}$ is the constitutive stiffness matrix.
\subsubsection{Concentrated gravitational potential.}
Let $m_i$ denote the lumped mass associated with disk $i+1$ (end of segment $i$).
The gravitational potential energy is
\begin{equation}
\label{eq:gravity_lumped}
\Pi_{\text{grav}}(\mathbf{q})
=
\sum_{i=1}^n m_i g\,\mathbf{e}_z^\top \mathbf{p}_{i+1},
\end{equation}
where $g$ is the gravitational acceleration with sign consistent along the chosen z-axis $\mathbf{e}_z$.

\subsubsection{Distributed backbone weight.}
Let $\mu$ be the effective mass per unit length of the backbone.
Approximating the distributed weight on each segment using midpoint quadrature gives
\begin{equation}
\label{eq:gravity_distributed}
\Pi_{\text{bb}}(\mathbf{q})
\approx
\sum_{i=1}^n \mu g\,L_i\,\mathbf{e}_z^\top \mathbf{p}_i(L_i/2),
\end{equation}
where $\mathbf{p}_i(L_i/2)$ is the backbone centerline position evaluated at the segment midpoint.

\subsubsection{Tendon contributions}
\label{tendonenergysection}
If the tendon is assumed to be inextensible \cite{rucker2011, rao2021model}, its length can be enforced as an equality constraint in the energy formulation using a Lagrange multiplier. Let the undeformed tendon length be $L_0$, and let the applied actuation (tendon pull) be $L_{\text{act}}$. At equilibrium, the total geometric length of the tendon plus the actuation satisfies $\sum_i L_{t,\text{seg}}^{(i)} + L_{\text{act}} = L_0$, where the tendon segment between two adjacent spacer disks is perfectly straight and is computed as $L_{t,\text{seg}}^{(i)} = \|\mathbf{x}_{i+1} - \mathbf{x}_i\|$. This constraint can be enforced in the energy formulation using a Lagrange multiplier as:
$\Pi_{\text{tend}}(\mathbf{q},\lambda) = \lambda \left( L_0 - \sum_i L_{t,\text{seg}}^{(i)} - L_{\text{act}} \right)$.
Assuming the tendon carries constant tension along its length, the Lagrange multiplier corresponds to the tendon tension, i.e., $\lambda = T$.
%But as discussed previously, we accommodate the elasticity of the tendon in this model using a linear spring approximation (\cref{model}). 

However, as discussed previously, we accommodate the elasticity of the tendon in this model using a linear spring approximation (\cref{model}) instead of enforcing an inextensibility constraint. Let the total routed tendon length over all segments $i$ be $L_{\text{route}} = \sum_i L_{t,\text{seg}}^{(i)}$.
The elastic extension of the tendon is then $\Delta L = L_{\text{route}} - (L_0 - L_{\text{act}})$. Using a linear spring approximation with stiffness $k_{ten}$, the tendon elastic potential energy is written as
\begin{equation}
\label{tendonenergy}
\Pi_{\text{tendon}}(\mathbf{q})
=
\frac{1}{2} k_{ten} \left( L_{\text{route}} - (L_0 - L_{\text{act}}) \right)^2.
\end{equation}

The linear spring stiffness can be modeled from the tendon's Young's modulus as: $k_{ten} = E_tA/L_0$, where $E_t$ and $A$ denote the tendon's elastic modulus and area respectively.

The total potential energy due to all contributing effects from \cref{eq:bending_energy}--\cref{tendonenergy} is given by
\begin{equation}
\label{total energy}
\Pi(\mathbf{q})
=
\Pi_{\mathrm{EP}}(\mathbf{q})
+
\Pi_{\mathrm{grav}}(\mathbf{q})
+
\Pi_{\mathrm{bb}}(\mathbf{q})
+
\Pi_{\mathrm{tendon}}(\mathbf{q}).
\end{equation}
The static equilibrium is obtained by solving the first-order optimality conditions, 
\begin{equation}
\label{eq:equilibrium_condition}
\nabla_{\mathbf{q}} \Pi(\mathbf{q}) = \mathbf{0}.
\end{equation}
With the exact analytical single-segment solutions and the sensitivities propagated as discussed in \cref{multiseg}, the PCS achieves fast numerical convergence using the Levenberg--Marquardt method, implemented within the \texttt{fsolve} framework in \textsc{MATLAB}~R2023a.

\subsection{Model Benchmarking}
We validate the formulated PCS model against a variable curvature model based on Cosserat Rod Theory
\cite{antman2005nonlinear},\cite{altenbach2013cosserat}. The local disk rotations require a modeling framework which incorporates the effects of a partially constrained tendon path. That is, the tendon exerts discrete forces and moments on the backbone in each segment at the tendon-disk connections. The exact framework is presented in \cite{rao2021model}, which we modified to incorporate local disk rotations. 
The net force and moment acting on the backbone at the $j$th disk due to the tendon are given by:
\[
\mathbf{F}_{j}
=
T_{j}
\frac{\overrightarrow{\mathbf{x}_{j}\mathbf{x}_{j-1}}}
{\left\|\overrightarrow{\mathbf{x}_{j}\mathbf{x}_{j-1}}\right\|}
+
T_{j+1}
\frac{\overrightarrow{\mathbf{x}_{j}\mathbf{x}_{j+1}}}
{\left\|\overrightarrow{\mathbf{x}_{j}\mathbf{x}_{j+1}}\right\|},
\]
\[
\mathbf{M}_{j}
=
\mathbf{r}(\alpha_j)
\times
\mathbf{F}_{j},
\]
where the tensions $T_j$ = $T_{j+1}$ when friction is neglected, and $\mathbf{r}(\alpha_j)$  is the tendon radial offset vector in the local disk frame after rotation $\alpha$.

For an $n$-segment Cosserat rod in static equilibrium, the internal force $\mathbf{n}(s)$ and internal moment $\mathbf{m}(s)$ satisfy the balance relations:
\[
\mathbf{n}'(s) + \sum_{k=1}^{n} \mathbf{f}_k(s) = 0,
\]
\vspace{-4mm}
\[
\mathbf{m}'(s) + \mathbf{p}'(s) \times \mathbf{n}(s) + \sum_{k=1}^{n} \mathbf{l}_k(s) = 0,
\]
where $\mathbf{f}(s)$ and $\mathbf{l}(s)$ denote externally distributed force and moment densities. Gravitational loading is incorporated through $\mathbf{f}(s)$, and the constitutive relations couple the strain measures to the internal force and moment resultants. The tendon termination loads for each segment are imposed via boundary (jump) conditions. The resulting discontinuities in internal force and moment at a segment end $\ell$ are given by: 
\[
\mathbf{n}(\ell^-) = \mathbf{n}(\ell^+) + \sum_{k=1}^{n} \mathbf{F}_k + \mathbf{F}_{\text{ext}},\]
\vspace{-3mm}
\[
\mathbf{m}(\ell^-) = \mathbf{m}(\ell^+) + \sum_{k=1}^{n} \mathbf{M}_k.
\]
The VC model does not incorporate tendon elasticity nor spacer disk thickness. In addition, the VC model is tension-controlled, whereas the PCS model is displacement-controlled. In the PCS model without tendon elongation considerations, the constant tension along the tendon can be obtained from the Lagrange multiplier to the inextensibility constraint as was discussed in \cref{tendonenergysection}. Thus, for a commanded tendon displacement, we first solve the PCS model for the Lagrange multiplier and use it as the tendon tension in the VC model. Moreover, since the PCS model incorporates backbone weight as a midpoint quadrature formulation compared to the distributed loading method in the VC model, we do not consider backbone weight for this model comparison. The geometric properties of the MDL manipulator (\cref{tab:robot_params}) are considered. In addition, the backbone's Young's modulus is set at 60 GPa. This is an approximate estimate for Nitinol. The exact Young's modulus used for experimental validation is calibrated using experimental data from the manipulator prototype (\cref{validationsection}).

In order to compare both models, we evaluate multiple tendon routing scenarios: (a) parallel, (b) single disk rotated, and (c) more than one disk rotated. All evaluations were performed on a MacBook Pro equipped with an Apple M2 processor and 16GB of RAM. We evaluate the PCS model with respect to the VC model based on its accuracy and computational expense. Two error metrics were employed to evaluate the accuracy of the PCS model: the root mean square (RMSE) and tip error. \cref{tab:avg_performance} shows the average RMSE, tip errors, and computational time taken by both of the models for eleven different tendon actuations ranging from 0.5 - 3.5 cm in increments of 0.25 cm.
\begin{table}[]
\vspace{3.5mm}
\centering
\caption{Model Comparison}
\label{tab:avg_performance}
\setlength{\tabcolsep}{3pt}
\begin{tabular}{lcccc}
\toprule
\textbf{Routing} &
\makecell{\textbf{Avg. RMSE}\\(cm)} &
\makecell{\textbf{Avg. Tip Error}\\(cm)} &
\makecell{\textbf{Avg. Time}\\PCS (s)} &
\makecell{\textbf{Avg. Time}\\VC (s)} \\
\midrule
Parallel & 0.0144 & 0.0309 & 0.7925 & 7.65 \\
D6 $90^\circ$ CW & 0.0778 & 0.1476 & 0.7769 & 9.75 \\
\makecell[l]{D6 $90^\circ$ CW + \\ D9 $90^\circ$ CCW}
& 0.1737 & 0.3022 & 0.7761 & 15.7362 \\
\bottomrule
\end{tabular}
\end{table}
From \cref{tab:avg_performance}, it can be observed that the PCS model is superior in computational performance without compromising on the model accuracy. For the parallel case, an average tip error of 0.03 cm ($\sim$0.1 \%) is reported, while the PCS model performs $\sim$10 times faster. Similarly, the average tip error in the disk rotation cases are $\sim$0.5\% and $\sim$1.1\%, while the PCS solver is $\sim$20 times faster. It must be also noted that the VC model solves the Cosserat ODEs using a shooting method, which sometimes fails to converge and often struggles with numerical difficulties with increased number of disks and/or increased disk actuations. The PCS model, though, is more stable numerically due to analytical gradient computations for the PCS segment.

\section{Experimental Methods}
\label{validationsection}
The model presented in \cref{modelsection} is validated with results from an experimental environment; the smaller MDL manipulator is used for the experiments, and the larger CDR manipulator is manufactured as a proof-of-concept. Three sets of experiments are performed: (a) the tendon routed parallel to the backbone, (b) an intermediate disk (D6) rotated $90^\circ$ clockwise, and (c) two intermediate disks (D6 and D9) rotated $90^\circ$ clockwise and counterclockwise respectively. Each scenario is evaluated for tendon actuations from 1.0 to 3.0 cm in 0.5 cm increments. 

\subsection{Experimental Setup}
\label{setup}
\begin{figure}[h]
    \centering    
    \includegraphics[width=0.4\textwidth]{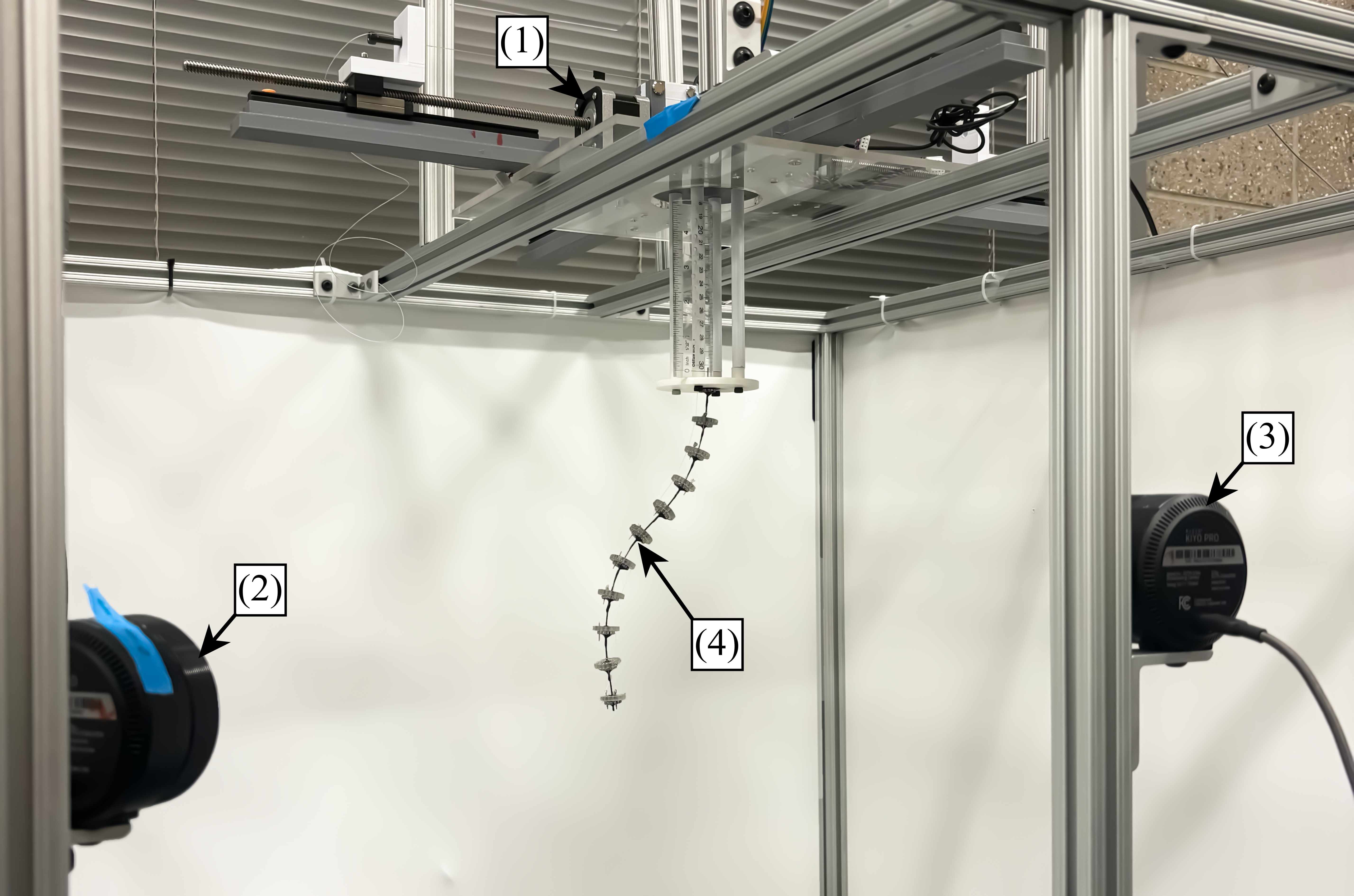} 
    \caption{The MDL manipulator (4) is suspended within an enclosure, with the tendon being actuated by the stepper motor (1). Two cameras (2,3) capture the manipulator's deformation by reconstructing the spatial coordinates of the spacer disks.}
    % Experimental setup with stepper motor (1) to actuate the tendon, and Camera 1 (2) Camera 2 (3) to retrieve track coordinates of the backbone. (4) is the manipulator under validation tests.}
    \label{fig:experiment_environment}
\end{figure}

% \begin{table}[htbp]
% \caption{Geometric Parameters of the Continuum Robot Prototypes}
% \label{tab:robot_params}
% \centering
% \begin{tabular}{lcccc}
% \toprule
% \textbf{Parameter} & \textbf{Symbol} & \textbf{Robot A} & \textbf{Robot B} & \textbf{Unit} \\
%  & & \textit{(Large)} & \textit{(Small)} & \\
% \midrule
% \multicolumn{5}{l}{\textit{Geometric Parameters}} \\
% Total Length & $L_0$ & 552 & 270 & mm \\
% Number of Segments & $N$ & 8 & 10 & - \\
% Segment Length & $L_{seg}$ & 69 & 27 & mm \\
% Backbone Diameter & $d_b$ & 1.5 & 1.0 & mm \\
% Tendon Radius & $r_{pitch}$ & 34 & 8 & mm \\
% Disk Radius & $r_{disk}$ & 40 & 10 & mm \\

% Single Disk Mass & $m_{disk}$ & 40 & 2 & g \\
% \bottomrule
% \end{tabular}
% \end{table}
As demonstrated in \cref{fig:experiment_environment}, the MDL manipulator (4) is suspended under gravity within an enclosure built using 80/20 aluminum extrusions. The tendon is actuated linearly using NEMA-17 stepper motors (1) and the tendon pull commands are directly incorporated in the displacement controlled PCS model. Two RGB monocular cameras (Razer Kiyo Pro) with resolution 1920x1080 are used to reconstruct the spatial coordinates of the spacer disks. The cameras have identical intrinsic parameters, are calibrated relative to a fixed point in 3D, and capture the manipulator's pose from two different angles at a frequency of 30 Hz. In order to avoid occlusion and color discrepancies, the centers of the spacer disks are annotated manually, which are then triangulated to obtain the spatial coordinates with respect to a fixed frame. In order to compare with the model, another transformation matrix maps the spatial coordinates to the base of the manipulator. The manipulator's deformed configurations are captured for each of the three routing scenarios discussed previously. To avoid experimental errors, repeatability is taken into account by performing each experiment three times and taking the average.  
% We validated our model using a small robot we built within a larger framework.(\cref{fig:experiment_environment}) We use a computer with an Intel(R) Core(TM) i7-10700 CPU to control and perceive the robot. To control them, we employed a NAMA 17 stepper motor (\cref{fig:experiment_environment}-(1)). The motors were mounted on a fixed support above the robot(\cref{fig:experiment_environment}-(4)).To get the data of disks' positions, we constructed a vision-based positioning system using two Razer Kiyo Pro cameras with a resolution of 1920x1080 (\cref{fig:experiment_environment}-(2)(3)). Two cameras observe from different angles at a frequency of 30Hz, and the spatial position of the disc is obtained through manual annotation and triangulation calculation after calibration.To reduce human error, each experimental data point was measured independently three times, and the average value was used for subsequent experiments.
% \label{setup}

\subsection{Parameter calibration}
\label{calib}
% \note{GK: we need to clearly mention that the parameters are optimized for the parallel tendon orientation, and then applied to the disk rotation case. Consider the narrative commented below. Also is the tendon elasticity not a parameter?}
% There are multiple experimental parameters in the manipulator which need to be calibrated using the experimental ground truth. Namely, the addition of spacer disks, along with the binding epoxy layer increases the overall flexural rigidity of the backbone and changes its overall flexible length. Since precise measurement of the variations in these parameters is difficult, we implement a simple optimization framework to minimize the error of the model from the experimental ground truth. We use experimental results from five datasets with a parallel routed tendon, then calibrate $EI$ and $t_\text{rigid}$ from those results to be used for all other experiments. The error metric is defined as a squared sum of the discrete curvature error (capturing shape) and tip error. The error minimization yields parameters of $EI = 9.1579 \times 10^{-3} \, \text{N}\cdot\text{m}^2$, and $t_\text{rigid} = 2.3 \text{mm}$.  

There are multiple model parameters that need to be calibrated using the experimental ground truth for model validation. These are (a) $t_{rigid}$: the parts of the backbone that deform as a rigid body due to a relatively thick disk bonded rigidly to the backbone, and (b) $EI$: flexural rigidity of the backbone. While the diameter of the backbone wire is accurately measured, its Young's modulus is not. Furthermore, the addition of the epoxy to bond the disks on the backbone may change its flexural rigidity. We calibrate these parameters using a simple optimization framework to minimize the error between the model and experimental ground truth from the five datasets of the parallel routed case. Once the parameter values are identified from that data, the model is prepared to predict the deformation shapes when disks are rotated. The calibration error metric is defined as a squared sum of the discrete curvature error (capturing shape) and tip error. The error minimization yields parameters of $EI = 9.1579 \times 10^{-3} \, \text{N}\cdot\text{m}^2$, and $t_\text{rigid} = 2.3 \text{mm}$.  
\section{Results and Discussion}
\begin{figure*}[!t]
    \centering    \includegraphics[width=0.87\textwidth]{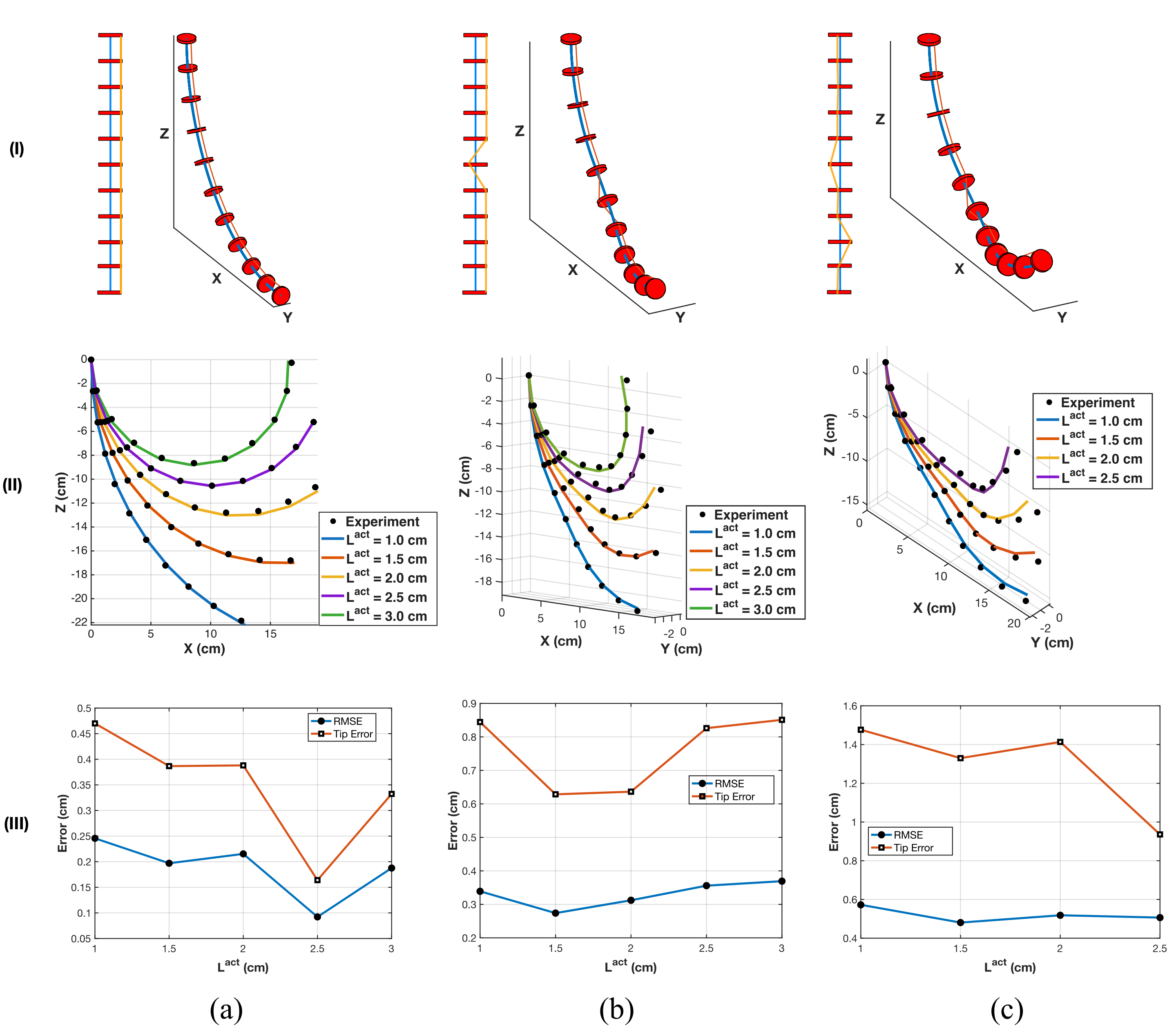} 
    \caption{The PCS model is validated against experimental results in three different tendon routing scenarios (a)-I: parallel, (b)-I: disk 6 rotated $90^\circ$ CW, and (c)-I: disk 6 and 9 rotated $90^\circ$ in opposite directions. The second row (II) demonstrates the model deformations for different tendon actuation using smooth curves, compared with black filled circles denoting experimental data, for each of the three routing scenarios (a)-(c). The RMSE and tip errors are plotted for each routing scenario in row (III).}
    \label{validation}
\end{figure*}
This section demonstrates the experimental validation of the predictive model developed in \cref{modelsection}. The PCS model is compared against experimental data from the three different tendon routing scenarios discussed in \cref{validationsection}. For each of the commanded tendon pulls and disk rotations, the model considers the optimal flexural rigidity and rigid path thickness obtained from the parameter optimization. In addition, we consider an elastic modulus $E_t = 4.5 \,\text{GPa}$ for the high-strength Nylon monofilament fishing line (diameter 0.45 mm) used as the tendon. This is consistent with reported ranges of 3-5 GPa for Nylon strings \cite{lynch2017mechanical}. Since drawn Nylon monofilaments typically have higher stiffness than bulk Nylon due to molecular alignment during drawing, we consider a value towards the upper end of this range. The equivalent spring stiffness ($k_{ten}$) is then evaluated as $k_{ten} = E_tA/L_0$ and incorporated in the tendon energy contribution (\cref{tendonenergysection}).
The experimental data obtained in \cref{validationsection} is compared with the results from the predictive model using two error metrics: (a) RMSE between disk center positions, and (b) tip errors. 
For the parallel and single disk rotation scenarios, the tendon is actuated from 1.0 to 3.0 cm in steps of 0.5 cm, and the spatial disk coordinates are captured using the two cameras. For the two-disk rotation case, actuation was limited to 2.5 cm, beyond which the deformed manipulator moved out of the camera’s field of view.
\cref{validation} shows the results for all scenarios.
In \cref{validation}(a), the tendon is routed parallel to the backbone, as shown in (I). \cref{validation}(a)-(II) shows the deformed configurations from the model (colored smooth curves), overlaid with the experimental points (black filled circles), for each of the five commanded tendon actuations (1.0 - 3.0 cm, in steps of 0.5 cm). This baseline experiment was used for calibration of the backbone parameters i.e flexural rigidity (EI) and rigid part thickness ($t_{rigid}$). The RMSE and tip errors are plotted as a function of actuated length in \cref{validation}(a)-(III), which are reported to be under 0.25 cm ($\sim$0.9\% of manipulator length) and 0.47 cm ($\sim$1.7\%) respectively. The average tip error for all five experiments is $\sim$0.34 cm, corresponding to 1.2\% of the manipulator's length. This is consistent with \cite{rucker2011}, who report an average error of 1.5\% for their straight tendon routed experiment. Although the error magnitudes are similar for all tendon actuations, the slightly higher error at 1 cm tendon actuation may be attributed to the slight tendon slack before the start of the experiment and a small initial precurvature. 
Next, \cref{validation}(b) and (c) demonstrate the cases with disk rotations. In (b), disk 6 is rotated clockwise by $90^\circ$, which creates a torsion in the backbone path as shown in \cref{validation}(b)-I. The observed errors are higher than the parallel case, with RMSE $\le$ 0.37 cm ($\sim$1.1\%) and tip error $\le$ 0.85 cm ($\sim$3.1\%). The average tip error is $\sim$0.75 cm, corresponding to $\sim$2.8\% of the manipulator's length. Though such a discrete tendon routing scenario has not been reported in literature, the error percentages are slightly higher than polynomial ($\sim$1.9\%) and helical ($\sim$2.27\%) routing reported in \cite{rucker2011}. Other works considering helical routings \cite{10606060} report an average error of $\sim$3.1\% which is still consistent with our results. The increase in error can be attributed to friction effects between the nylon tendons and the disk holes through which they are routed. Similar effects were observed with the two disk rotation case, as demonstrated in \cref{validation}(c)-III, where the maximum tip error reported is around 1.44 cm ($\sim$5.3\%), while the maximum RMSE is around 0.57 cm ($\sim$2.1\%). The average tip error is around 1.03 cm, corresponding to around 3.5\% of the manipulator's length. 

\section{Conclusion and Future Work}
This work demonstrated the design and modeling of a novel tendon driven continuum manipulator capable of locally rerouting tendons by rotation of the intermediate spacer disks. Two designs were presented: (a) the MDL, where the disks can be manually rotated and locked in place, and (b) the CDR, where the disks can actively rotate either prior to or during operation. A Piecewise Constant Strain statics model was developed from a potential energy formulation by taking into consideration: (a) disk rotations and discrete linear tendon paths between rotated disks, (b) reduction in effective bending length of the backbone due to the presence of rigid spacer disks, and (c) axial elastic elongation of the tendon during actuation. The developed PCS model was compared with a higher-fidelity Variable Curvature Cosserat Rod model and was demonstrated to have sufficient accuracy with significantly lower computational expense ($\sim$ one order magnitude lower), making it a suitable modeling framework for future efforts in design optimization and controls. Implementing workspace constraints, end-effector reachability targets, and inverse kinematics can be simplified by incorporating these into the energy formulation. 
\begin{figure}[h]
    \centering    
    \includegraphics[width=0.43\textwidth]{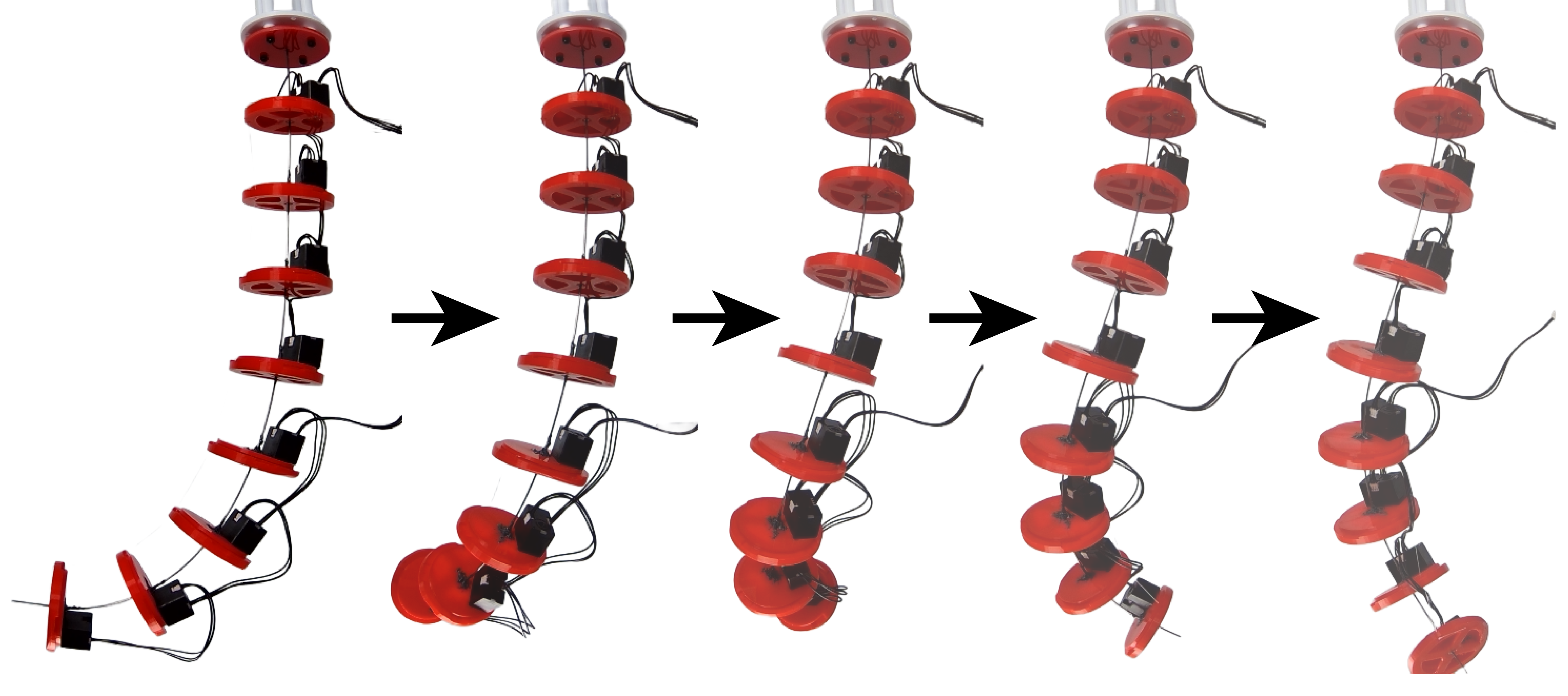} 
    \caption{CDR manipulator demonstrating tendon rerouting \textit{on-the-fly} achieving a helical deformation profile from an initial parallel configuration}
    \label{bigmanipulator}
\end{figure}

This paper also focused on validation of the smaller MDL manipulator and manufacturing the CDR manipulator as a proof-of-concept. Our previous work \cite{11522832} demonstrated a simplified actuation strategy for shape matching with the CDR manipulator. Future efforts will focus on demonstrating real-world robotic applications using active spacer disk rotation of the CDR manipulator.  \cref{bigmanipulator} demonstrates one such experimental scenario, where the CDR manipulator achieves a helical orientation from an initial parallel configuration exclusively by selective disk rotations. Future modifications to the PCS model will involve incorporating the backbone's precurvature and the tendon's slack parameters. Though negligible, these effects may lead to discrepancies in manipulators with larger number of segments. It was also observed that the accuracy of the model reduced with increasing disk rotation; this result can be attributed to the friction between the tendons and the disk guide holes, which gets amplified with greater angles of disk rotation. Incorporating the non-conservative friction effects into a variational energy formulation is not straightforward and typically requires a dissipation model within a quasi-static or dynamics framework. Future efforts will explore embedding such effects within the energy framework. In addition, we will conduct a comprehensive workspace and dexterity evaluation of the proposed RTDCM design to identify redundant disks that do not contribute as much to the manipulator's performance, thereby informing and guiding design optimization.  
\bibliographystyle{IEEEtran}
\bibliography{references}
\end{document}